\documentclass[lettersize,journal]{IEEEtran}
\usepackage{amsmath,amssymb,amsfonts}
\usepackage{algorithm}
\usepackage{algorithmic}
\usepackage{graphicx}
\usepackage{multirow}
\usepackage{array}
\usepackage{subfigure}
\usepackage{tabularx}
\usepackage{siunitx}
\usepackage{textcomp}
\usepackage{adjustbox} 
\usepackage{colortbl} 
\usepackage{xcolor}
\usepackage{booktabs}
\usepackage{tcolorbox}
\usepackage[colorlinks,linkcolor=blue,anchorcolor=blue, urlcolor=blue, citecolor=blue]{hyperref}

\usepackage{pifont}

\title{Pentest-R1: Towards Autonomous Penetration Testing Reasoning Optimized via Two‑Stage Reinforcement Learning}

\usepackage{bibentry}
\usepackage{pifont}

\begin{document}

\author{
  \rm He Kong\textsuperscript{1,2}, 
  \rm Die Hu\textsuperscript{1,2}, 
  \rm Jingguo Ge\textsuperscript{1,2}, 
  \rm Liangxiong Li\textsuperscript{1}, 
  \rm Hui Li\textsuperscript{1} , and 
  \rm Tong Li\textsuperscript{1} \\
  \textsuperscript{1}State Key Laboratory of Cyberspace Security Defense, Institute of Information Engineering, \\Chinese Academy of Sciences \\
  \textsuperscript{2}School of Cyber Security, University of Chinese Academy of Sciences
}
\maketitle

\begin{abstract}
Automating penetration testing is crucial for enhancing cybersecurity, yet current Large Language Models (LLMs) face significant limitations in this domain, including poor error handling, inefficient reasoning, and an inability to perform complex end-to-end tasks autonomously. To address these challenges, we introduce Pentest-R1, a novel framework designed to optimize LLM reasoning capabilities for this task through a two-stage reinforcement learning pipeline. We first construct a dataset of over 500 real-world, multi-step walkthroughs, which Pentest-R1 leverages for offline reinforcement learning (RL) to instill foundational attack logic. Subsequently, the LLM is fine-tuned via online RL in an interactive Capture The Flag (CTF) environment, where it learns directly from environmental feedback to develop robust error self-correction and adaptive strategies. Our extensive experiments on the Cybench and AutoPenBench benchmarks demonstrate the framework's effectiveness. On AutoPenBench, Pentest-R1 achieves a 24.2\% success rate, surpassing most state-of-the-art models and ranking second only to Gemini 2.5 Flash. On Cybench, it attains a 15.0\% success rate in unguided tasks, establishing a new state-of-the-art for open-source LLMs and matching the performance of top proprietary models. Ablation studies confirm that the synergy of both training stages is critical to its success. The source code and datasets are publicly accessible in the \url{https://github.com/KHenryAegis/Pentest-R1}.
\end{abstract}

\begin{figure*}[t!]
\centering
\includegraphics[width=\linewidth]{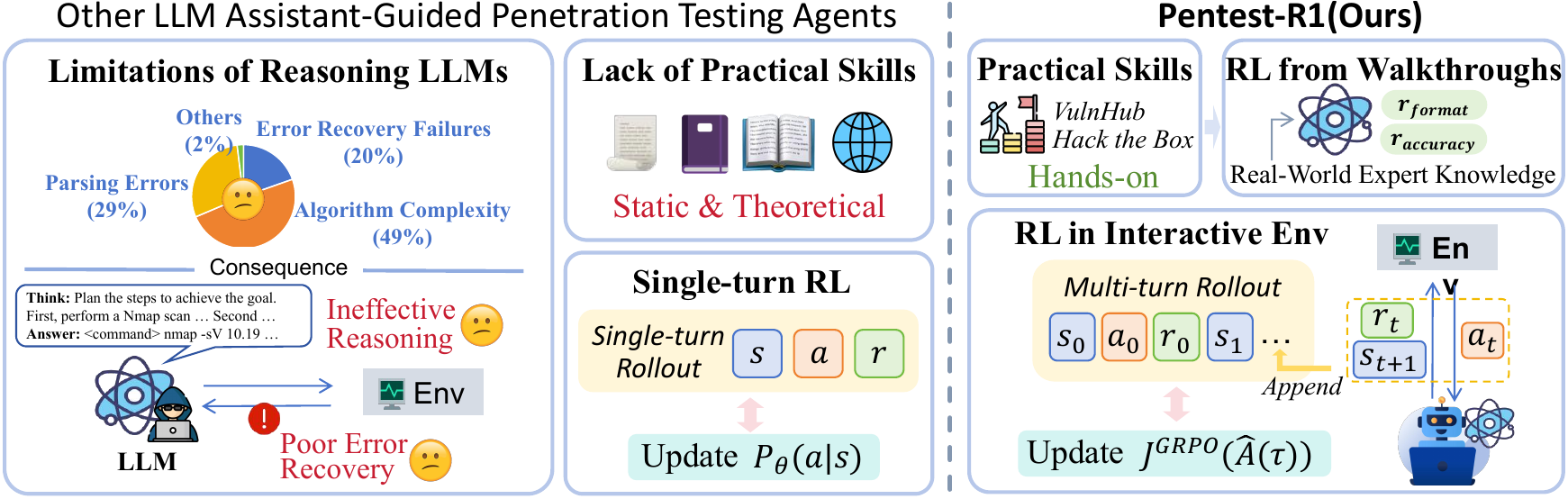}
\caption{An illustration of the core challenges in LLM-driven penetration testing. Current LLMs often exhibit ineffective reasoning and poor error recovery, leading to failed attack chains. This is exacerbated by a lack of realistic, multi-step training data and RL paradigms ill-suited for long-horizon, interactive tasks.}
\label{fig:motivation}
\end{figure*}

\section{Introduction}
Penetration testing is a proactive and indispensable technique for identifying, assessing, and mitigating security vulnerabilities. However, traditional workflows are heavily reliant on the deep expertise and significant time investment of human professionals \cite{metasploit2020}.
The recent advancement of Large Language Models (LLMs) \cite{achiam2023gpt,guo2025deepseek,comanici2025gemini} has introduced a promising new frontier for automating penetration testing, offering the potential to improve efficiency, reduce costs, and broaden the scope of security assessments \cite{deng2024pentestgpt,kong2025vulnbot}.
Despite this promise, current LLMs struggle to execute complex, end-to-end penetration testing tasks autonomously. While prompting strategies like Chain-of-Thought (CoT) \cite{wei2022chain} can enhance reasoning, they often produce verbose thought processes that incur significant computational overhead and are ill-suited for the rapid, interactive nature of security engagements \cite{yang2023intercode}. Furthermore, existing reinforcement learning (RL) applications are often designed for single-turn, sparse-reward scenarios, a paradigm that fails to capture the multi-round, stochastic, and strategically complex nature of penetration testing \cite{foley2025apirl,xiong2025minimalist, mroueh2025reinforcement}. This mismatch, compounded by a lack of large-scale, multi-step datasets from real-world scenarios, has hindered the development of truly autonomous agents capable of sophisticated attack planning and execution.

To address these challenges, we introduce Pentest-R1, a novel framework that enhances and optimizes the reasoning capabilities of LLMs through a specialized two-stage reinforcement learning pipeline. The first stage, \textit{offline RL}, instills foundational knowledge by training the LLMs on a new, curated dataset of over 500 real-world expert walkthroughs from platforms like HackTheBox and VulnHub. This dataset, structured in a unique "Thought-Command-Observation" format, captures the complete sequence of an expert's cognitive process, command execution, and resulting outcomes. In the second stage, \textit{online RL}, the pre-trained LLMs are fine-tuned within an interactive Capture The Flag (CTF) environment. Here, it learns directly from environmental feedback, refining its policy through trial and error to develop robust error self-correction and adaptive strategies. To ensure efficient and stable training, the framework leverages Group Relative Policy Optimization (GRPO) \cite{shao2024deepseekmath}, a critic-free algorithm, in conjunction with Low-Rank Adaptation (LoRA) \cite{hu2022lora}.
We conducted extensive experiments on the challenging Cybench \cite{zhang2024cybench} and AutoPenBench \cite{gioacchini2024autopenbench} benchmarks demonstrate the framework's effectiveness. 
Our results demonstrate that Pentest-R1 achieves a 24.2\% success rate on AutoPenBench, outperforming the majority of state-of-the-art (SOTA) models, including GPT-4o, and ranking second only to Gemini 2.5 Flash. On Cybench, it achieves a SOTA score of 15.0\% in unguided tasks, placing it in a tie for first place overall among all tested models. 
Ablation studies confirm that the synergy of both offline knowledge acquisition and online interactive refinement is critical to its success. This work presents the first framework to successfully employ end-to-end online reinforcement learning for penetration testing, bridging the gap between static knowledge and dynamic, real-world applications.

The contributions of this work are as follows:
\begin{enumerate}
    \item We construct a large-scale dataset of multi-step penetration testing walkthroughs. Compiled from more than 500 real-world scenarios on platforms like HackTheBox and VulnHub, this dataset is structured into "Thought-Command-Observation" tuples. It provides an unprecedented resource for training agents to understand complex attack logic.
    \item We propose the first two-stage, end-to-end reinforcement learning framework for autonomous penetration testing. Pentest-R1 uniquely combines offline RL on expert data with online RL in an interactive environment. This design not only imparts foundational domain knowledge but, crucially, enables the LLMs to learn adaptive strategies and robust error self-correction capabilities from direct interaction and feedback.
    \item We demonstrate state-of-the-art performance through rigorous experimental validation. Our evaluation on the Cybench and AutoPenBench benchmarks shows that Pentest-R1 significantly outperforms other open-source models and achieves performance comparable to proprietary SOTA models.
\end{enumerate}
\section{Related Work}
\subsection{Penetration Testing}
Traditional automated penetration testing systems rely on rule-based engines (e.g., Metasploit \cite{metasploit2020} and AutoSploit \cite{moscovich2020autosploit}) that use hard-coded attack patterns, limiting their adaptability to dynamic environments.
Recent advancements address these limitations by integrating LLMs and multi-agent frameworks \cite{happe2025can, mayoral2025cai}. For example, PentestGPT \cite{deng2024pentestgpt} uses hierarchical task trees to guide exploitation, while VulnBot \cite{kong2025vulnbot} leverages role-specialized collaboration to streamline penetration workflows. AUTOATTACKER \cite{xu2024autoattacker} further demonstrates the potential of LLMs in automating multi-stage attacks. Benchmarks such as Cybench \cite{zhang2024cybench}, AUTOPENBENCH \cite{gioacchini2024autopenbench}, and NYU CTF Bench \cite{NEURIPS2024_69d97a64} help standardize evaluations across varied attack scenarios, highlighting LLMs’ strengths in task decomposition. However, persistent challenges remain—particularly in real-time interaction with dynamic environments \cite{yang2023intercode}.
The automation of penetration testing has advanced significantly, yet critical gaps remain in end-to-end penetration testing. While large language models (LLMs) show promise, they often generate invalid tool invocations, hallucinate parameters, or fail to adapt to dynamic network environments. Current systems lack robust multi-step reasoning and real-time situational awareness, limiting their ability to adjust commands based on live feedback or emerging attack surfaces.

\subsection{Large Language Models}
The rapid advancement of LLMs has significantly improved their performance in complex reasoning tasks, positioning them as valuable tools for high-precision domains like competitive programming \cite{el2025competitive}. Foundational models such as OpenAI’s o1 \cite{jaech2024openai} and GPT-4o \cite{hurst2024gpt} introduced systematic reasoning methodologies, inspiring derivative models like Llama-berry \cite{zhang2024llama}, Llava-o1 \cite{xu2024llava}, o1-Coder \cite{zhang2024o1}, and Marco-o1 \cite{zhao2024marco}, which extend reasoning capabilities to mathematical problem-solving and code generation. Meanwhile, models like DeepSeek-R1 \cite{guo2025deepseek}, QwQ \cite{qwq32b}, and Gemini \cite{team2023gemini} demonstrate strong reasoning performance, underscoring the diverse potential of LLMs in complex tasks. Despite these advancements, their applicability remains constrained in highly specialized areas such as penetration testing.
Existing studies have combined LLM-based agents with reinforcement learning for reasoning and planning tasks \cite{deng2024novice,feng2025group}. For example, Agent‑R1 \cite{Agent-R1} proposes a framework that employs end-to-end reinforcement learning to train LLM agents over multiple turns, thereby automating the handling of complex tasks. Similarly, Search‑R1 \cite{jin2025search} uses reinforcement learning to train an LLM to autonomously generate multi-turn search queries during stepwise reasoning, optimizing the inference trajectory. In addition, works such as RAGEN \cite{wang2025ragen} have introduced multi-turn trajectory optimization methods tailored to LLM agents, emphasizing the importance of fine-grained reward signals. DoctorAgent‑RL \cite{feng2025doctoragent}, a multi-agent cooperative reinforcement learning system for medical dialogue, models the consultation process as a collaborative multi-agent RL framework, achieving clinician-like performance in AI-driven medical questioning.
\section{Motivation}
\textbf{Are current reasoning models effective for automated penetration testing?} As the complexity of cybersecurity threats continues to grow, the need for automated solutions to rigorously assess system security has become increasingly urgent. While traditional penetration testing is a labor-intensive process requiring deep expertise, the rise of LLMs presents a promising avenue for automation. However, a critical question remains: whether today's LLMs are truly prepared for the nuanced, multi-step reasoning required for effective penetration testing remains an open question.

To explore this question, we conducted a preliminary study evaluating three leading LLMs (Qwen3-32B \cite{yang2025qwen3}, Claude 3.7 Sonnet \cite{claude3.7sonnet}, and Gemini 2.5 Flash \cite{comanici2025gemini}) on the Cybench and AutoPenBench benchmarks. We tested each model in two configurations: (1) \textit{with thinking}, leveraging native Chain-of-Thought (CoT) capabilities, and (2) \textit{without thinking}, using a direct-response mode. The agents operated autonomously on a Kali Linux host \cite{kali}, with performance measured by the \texttt{pass@3} success rate (i.e., capturing the flag in at least one of three independent trials). Each model is allotted 15 interaction turns on Cybench and 30 turns on AutoPenBench. 

\begin{table}[h]
\centering
\caption{Performance (\texttt{pass@3}) of state-of-the-art LLMs on autonomous penetration testing tasks. Configurations: w/t = with chain-of-thought reasoning, w/o t = without.}
\label{tab:llm_motivation_results}
\resizebox{0.5\textwidth}{!}{%
\begin{tabular}{@{}lcc@{}}
\toprule
\textbf{Model (Configuration)} & \textbf{Cybench (\%)} & \textbf{AutoPen (\%)} \\
\midrule
Qwen3-32B (w/t) & 5.0 & 9.1 \\
Qwen3-32B (w/o t) & 5.0 & 12.1 \\
\addlinespace
Claude 3.7 Sonnet (w/t) & 15.0 & 9.1 \\
Claude 3.7 Sonnet (w/o t) & 15.0 & 15.2 \\
\addlinespace
Gemini 2.5 Flash (w/t) & 10.0 & 27.3 \\
Gemini 2.5 Flash (w/o t) & 10.0 & 18.2 \\
\bottomrule
\end{tabular}
}
\end{table}

The results, summarized in Table~\ref{tab:llm_motivation_results}, reveal that chain-of-thought (\textit{with thinking}) does not guarantee improved performance and can even be detrimental.  These performance gaps stem from more fundamental weaknesses. As depicted in Figure~\ref{fig:motivation}, LLMs frequently fail due to an inability to recover from execution errors, handle the algorithmic complexity of a multi-step plan, or correctly parse feedback from the environment. These specific failures point to two systemic challenges that our work aims to solve:

\textbf{Challenge 1: Lack of Multi-Step Walkthrough Data in Real-world Environments.}
A major impediment to progress is the scarcity of high-quality, multi-step datasets that reflect real-world infiltration scenarios. Existing resources are often too simplistic or fragmented to train agents for complex, end-to-end penetration testing tasks.

\textbf{Challenge 2: The Single-Round Training Paradigm.}
Current reinforcement learning methods for LLMs are predominantly designed for single-turn, question-answering tasks. Penetration testing, however, is an inherently interactive, multi-round process characterized by stochastic feedback, where the outcome of one action unpredictably influences the next. This single-round training paradigm is fundamentally ill-suited for teaching the long-horizon, strategic reasoning required to navigate a dynamic attack surface.

\begin{figure*}[ht]
\centering
\includegraphics[width=\linewidth]{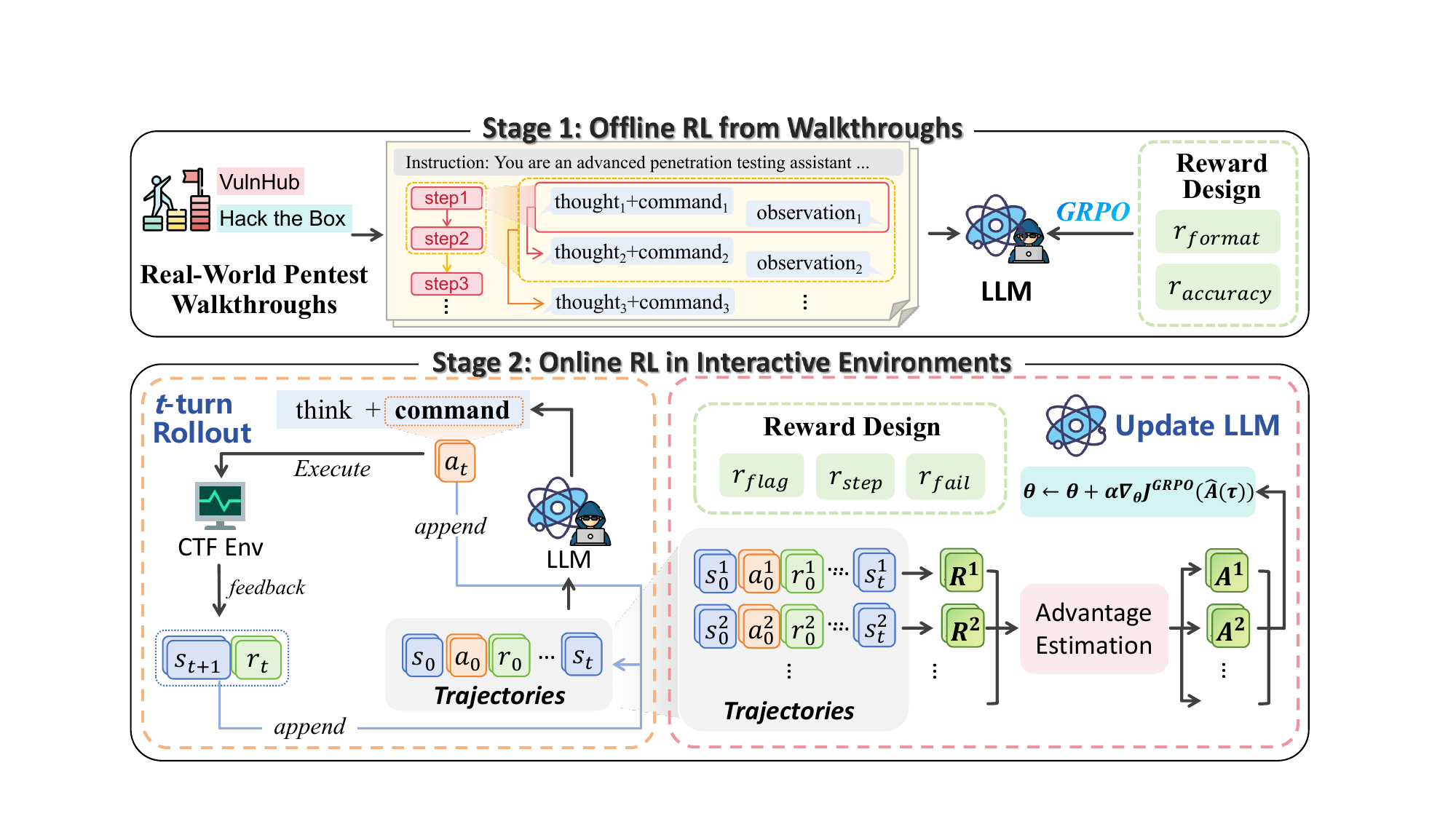}
\caption{The framework architecture of Pentest-R1.}
\label{fig:model}
\end{figure*}
\section{Methodology}

\subsection{Overview}
To overcome these challenges, we introduce Pentest-R1, a novel framework that leverages a two-stage reinforcement learning process, as shown in Figure \ref{fig:model}. Firstly, it learns domain knowledge from the real-world multi-step penetration test dataset. Secondly, it is explicitly designed to train LLM agents to learn from the rich feedback of an interactive environment, enabling them to build and refine complex penetration testing strategies autonomously.

\textbf{Policy Optimization with Group Relative Policy Optimization (GRPO).}
For both training stages, we employ Group Relative Policy Optimization (GRPO), a critic-free reinforcement learning algorithm. GRPO works by generating multiple candidate responses for a given state and evaluating them based on a reward function. The key idea is to compare each response not against a learned value function, but against the average performance of the other responses in its generation group. Maximizing this objective updates the policy's parameters $\theta$ to increase the probability of actions that yield above-average rewards.

While this core principle remains consistent, its application is tailored to the distinct structures of our two training stages. In Stage 1, GRPO optimizes single-turn responses on a static dataset. In Stage 2, it is extended to optimize entire multi-turn trajectories generated through live interaction with an environment.

\textbf{Low-Rank Adaptation (LoRA).}
To make the fine-tuning process computationally feasible, we utilize Low-Rank Adaptation (LoRA). Instead of updating all the weights of the LLM, LoRA freezes the pre-trained model and injects trainable, low-rank matrices into each layer. For a pre-trained weight matrix $W_0 \in \mathbb{R}^{d \times k}$, the update is represented by two low-rank matrices $A \in \mathbb{R}^{r \times k}$ and $B \in \mathbb{R}^{d \times r}$, where the rank $r \ll \min(d, k)$. The modified forward pass becomes:
\begin{equation}
h = W_0 x + \alpha B (A x)
\end{equation}
where $\alpha$ is a scaling factor, and only $A$ and $B$ are updated during training (with $r=32$ and $\alpha=64$ in our experiments). This dramatically reduces the number of trainable parameters, making fine-tuning faster and less memory-intensive without sacrificing performance.

\subsection{Stage 1: Offline Reinforcement Learning from Walkthroughs}
The initial training phase employs offline reinforcement learning to instill foundational penetration testing logic into the LLMs. This is achieved by fine-tuning the LLM on a curated dataset derived from over 500 expert demonstrations on platforms such as Hack The Box \cite{hackthebox} and VulnHub \cite{vulnhub}. Our data creation pipeline begins by collecting publicly available walkthroughs. From these, we programmatically extract the executed commands and their resulting observations. To reconstruct the expert's cognitive process, we employ an auxiliary LLM to reverse-engineer a plausible \textit{Thought} that would precede each command. Finally, these synthesized \textit{Thought-Command-Observation} tuples undergo manual proofreading to ensure high quality and logical consistency.
The training process is structured as a sequential, auto-regressive task designed to teach the model how to chain its reasoning based on evolving context. The model is conditioned on the entire history up to a given step and tasked with predicting the subsequent \textit{Thought} and \textit{Command}. For instance, to generate the first action, the model receives only the initial prompt. To generate the second action, its context is expanded to include the initial prompt plus the first \textit{Thought}, \textit{Command}, and the resulting \textit{Observation}. This iterative process continues for each step in the walkthrough. This methodology yields a final dataset of 14K high-quality, multi-turn interaction tuples. We then apply GRPO to this static dataset, treating each turn as a context-response pair, to fine-tune the agent's policy through offline reinforcement learning.

\paragraph{Reward Function.}
To guide the learning process, we define a composite reward function that evaluates the quality of an agent's generated completion $c$ against the ground-truth expert completion $c^*$.
\begin{equation}
R_{\text{offline}}(c, c^*) = R_{\text{format}}(c) + R_{\text{accuracy}}(c, c^*)
\end{equation}
The reward components are defined as follows:
\begin{itemize}
     \item \textbf{$R_{\text{format}}(c)$}: Encourages adherence to structure. A reward of +0.3 is given for correctly including the \texttt{<think>} block, and +0.3 for using the correct \textit{=== Step i ===} header, for a maximum of 0.6.
    \item \textbf{$R_{\text{accuracy}}(c, c^*)$}: Measures correctness. A reward of +0.2 is given for a matching step number. An exact command match yields +1.0. A partial match with Jaccard similarity $>$ 0.5 yields a scaled reward (e.g., $0.7 \times \text{Jaccard\_similarity}$).
\end{itemize}

\subsection{Stage 2: Online Reinforcement Learning in Interactive Environments}
In the second stage, the pre-trained agent is deployed into an interactive CTF environment, specifically InterCode-CTF \cite{yang2023intercode}, a benchmark for language-model-driven hacking tasks. Here, the agent learns from the consequences of its actions in real-time. The agent interacts with a sandboxed environment by issuing commands, receiving feedback (i.e., the next state) in the form of standard output or error messages. This trial-and-error process allows the agent to learn from its mistakes and reinforce successful strategies. In this stage, we use an enhanced version of GRPO with a loss mask, ensuring the loss is calculated only on tokens generated by the agent (assistant role) and not from the environment's feedback (user role).

\textbf{Episodic Trajectory Optimization with GRPO.}
Our approach in this stage enhances the standard GRPO algorithm for multi-turn, interactive tasks. Instead of scoring a single response, our agent generates an entire multi-step \textit{trajectory}, $\tau = (s_0, a_0, r_0, s_1, a_1, r_1, \dots, s_T)$, where $s_t$ represents the state (i.e., the cumulative observation history from the environment), $a_t$ is the agent's action (the \textit{Thought} and \textit{Command}), and $r_t$ is the immediate reward.
The final reward is \textit{episodic}, meaning a single scalar reward $R(\tau)$ is calculated based on the cumulative outcome of the entire trajectory (with shaped per-step rewards for valid actions and penalties for errors). This reward is then used to compute the advantage $\hat{A}(\tau)$. A crucial modification is the use of a turn-aware loss mask, $M$. This mask ensures that during backpropagation, the loss is calculated only on the tokens generated by the agent (the \textit{assistant} role) and not on the feedback from the environment (the \textit{user} role). The objective function for our multi-turn GRPO variant can be expressed as:
\begin{equation}
\begin{split}
\mathcal{L}_{\text{GRPO-MT}}(\theta) = -\mathbb{E}_{\tau \sim \pi_\theta} \Big[ \min\big(&\rho(\theta, \tau) \hat{A}(\tau), \\
&\text{clip}(\rho(\theta, \tau), 1-\epsilon, 1+\epsilon) \hat{A}(\tau)\big) \Big]
\end{split}
\end{equation}
where the policy ratio $\rho(\theta, \tau) = \frac{\pi_\theta(C_{asst}|\text{prompt})}{\pi_{\theta_{\text{old}}}(C_{asst}|\text{prompt})}$ (approximated per-token in practice for stability) is computed over $C_{asst}$, the concatenated sequence of all agent actions $(a_0, a_1, \dots)$ within the trajectory $\tau$, conditioned on the prompt and interleaved observations. The trajectory-level advantage $\hat{A}(\tau)$ is applied uniformly to all agent tokens via the loss mask $M$. This approach allows the model to learn complex, sequential decision-making from a holistic, trajectory-level reward signal. The specific algorithm is as shown in Algorithm \ref{alg:online_rl}.
\begin{algorithm}[H]
\caption{Online Trajectory Optimization in the Interactive Environment}
\label{alg:online_rl}
\begin{algorithmic}[1]
\STATE Initialize policy $\pi_\theta$ with weights from Stage 1.
\STATE Initialize CTF environment suite $\mathcal{E}$.
\STATE \textbf{for} each training epoch \textbf{do}
\STATE \quad \textbf{for} each task prompt $p$ from CTF dataset \textbf{do}
\STATE \quad \quad Initialize an empty list of trajectories $\mathcal{D} \leftarrow []$.
\STATE \quad \quad \textbf{for} $k=1$ to $N_{gen}$ (number of generations per prompt) \textbf{do}
\STATE \quad \quad \quad Reset environment $s_0 \leftarrow \mathcal{E}.\text{reset}(p)$.
\STATE \quad \quad \quad Initialize trajectory $\tau_k \leftarrow []$, cumulative reward $R_k \leftarrow 0$.
\STATE \quad \quad \quad \textbf{for} $t=0$ to $T_{max}$ (max steps) \textbf{do}
\STATE \quad \quad \quad \quad Generate response $y_t \sim \pi_\theta(s_t)$.
\STATE \quad \quad \quad \quad Extract action $a_t$ from response $y_t$.
\STATE \quad \quad \quad \quad Execute action in environment: $s_{t+1}, r_t, \text{done} \leftarrow \mathcal{E}.\text{step}(a_t)$.
\STATE \quad \quad \quad \quad Append $(s_t, y_t, a_t, r_t)$ to $\tau_k$.
\STATE \quad \quad \quad \quad $R_k \leftarrow R_k + r_t$.
\STATE \quad \quad \quad \quad \textbf{if} done \textbf{then break}
\STATE \quad \quad \quad \textbf{end for}
\STATE \quad \quad \quad Add $(\tau_k, R_k)$ to $\mathcal{D}$.
\STATE \quad \quad \textbf{end for}
\STATE \quad \quad Update policy $\theta \leftarrow \text{GRPO\_Update}(\theta, \mathcal{D})$ using rewards from $\mathcal{D}$.
\STATE \quad \textbf{end for}
\STATE \textbf{end for}
\end{algorithmic}
\end{algorithm}
\textbf{Reward Function.}
The reward signal in the online stage is designed to guide the agent towards successful task completion through effective exploration. The total episodic reward for a trajectory, $R_{\text{online}}(\tau)$, is the sum of rewards $r_t$ granted at each step $t$. The value of $r_t$ is determined by the outcome of the agent's action $a_t$ according to the following structure:
\begin{equation}
r_t = 
\begin{cases} 
r_{\text{flag}} & \text{if the flag is captured (task complete)} \\
r_{\text{step}} & \text{if action } a_t \text{ is valid and executes successfully} \\
r_{\text{fail}} & \text{if action } a_t \text{ is invalid or fails}
\end{cases}
\end{equation}
where the specific values for these components are defined to provide dense, incremental feedback:
\begin{itemize}
    \item \textbf{$r_{\text{flag}}$} is a large positive terminal reward, set to +1.0, awarded for successfully completing the task.
    \item \textbf{$r_{\text{step}}$} is an incremental reward of +0.1 (+0.05 for a valid command, +0.05 for successful execution) to encourage progress.
    \item \textbf{$r_{\text{fail}}$} represents a penalty of -0.1 for malformed or failed commands. A larger terminal penalty of -0.2 is applied if the episode ends with a failed flag submission.
\end{itemize}

\begin{table*}[h!]
\centering
\caption{Performance comparison of Pentest-R1 against state-of-the-art baseline models on AutoPenBench and Cybench. \textit{w/t} denotes with thinking (CoT); \textit{w/o t} denotes without. Llama3-70B’s 8K context is insufficient to complete the AutoPenBench. We evaluated the latest Claude 4 Opus on AutoPenBench.}
\label{tab:main_results}
\resizebox{\textwidth}{!}{%
\begin{tabular}{@{}lccccc@{}}
\toprule
\multirow{2}{*}{\textbf{Model}} & \multicolumn{2}{c}{\textbf{AutoPenBench}} & \multicolumn{3}{c}{\textbf{Cybench}} \\ \cmidrule(lr){2-3} \cmidrule(lr){4-6}
 & \textbf{Success Rate (\%)} & \textbf{Subtask-Success (\%)} & \textbf{Unguided (\%)} & \textbf{Subtask-Guided (\%)} & \textbf{Subtasks-Solved (\%)} \\ \midrule
GPT-4o & 21.2 & 29.0 & 12.5 & 17.5 & 28.7 \\
Claude 4 Opus / Claude 3 Opus & 18.2 & 16.4 & 10.0 & 12.5 & 36.8 \\
OpenAI o1-preview & 15.2 & 14.5 & 10.0 & 10.0 & 46.8 \\
Llama 3.1 405B Instruct & 6.1 & 7.9 & 7.5 & 15.0 & 20.5 \\
Mixtral 8x22b Instruct & 0.0 & 3.2 & 7.5 & 5.0 & 15.2 \\
Gemini 1.5 Pro & 9.1 & 22.1 & 7.5 & 5.0 & 11.7 \\
Llama 3 70b Chat & - & - & 5.0 & 7.5 & 8.2 \\
Qwen2.5-32B & 12.1 & 37.9 & 2.5 & 2.5 & 35.8 \\
Qwen3-32B (w/t) & 9.1 & 29.0 & 5.0 & 2.5 & 47.8 \\
Qwen3-32B (w/o t) & 12.1 & 36.9 & 5.0 & 7.5 & 44.3 \\
Claude 3.7 Sonnet (w/t) & 9.1 & 12.6 & 15.0 & 12.5 & 63.6 \\
Claude 3.7 Sonnet (w/o t) & 15.2 & 28.4 & 15.0 & 5.0 & 30.5 \\
Gemini 2.5 Flash (w/t) & 27.3 & 44.2 & 10.0 & 5.0 & 58.7 \\
Gemini 2.5 Flash (w/o t) & 18.2 & 39.1 & 10.0 & 5.0 & 45.6 \\
\midrule
\textbf{Pentest-R1 (Ours)} & \textbf{24.2} & \textbf{33.4} & \textbf{15.0} & \textbf{15.0} & \textbf{42.8} \\ \bottomrule
\end{tabular}%
}
\end{table*}

\begin{table*}[h!]
\centering
\caption{Ablation study of the Pentest-R1 framework. S1M, S1, and S2 denote Stage 1 Method, Stage 1, and Stage 2, respectively. \textit{SFT} and \textit{GRPO} are training methods. \checkmark indicates the component is included; $\times$ indicates it is not.}
\label{tab:ablation}
\resizebox{\textwidth}{!}{%
\begin{tabular}{@{}lccccccccc@{}}
\toprule
\multirow{2}{*}{\textbf{Model}} & \multicolumn{2}{c}{\textbf{S1M}} & \multirow{2}{*}{\textbf{S1}} & \multirow{2}{*}{\textbf{S2}} & \multicolumn{2}{c}{\textbf{AutoPenBench}} & \multicolumn{3}{c}{\textbf{Cybench}} \\
\cmidrule(lr){2-3} \cmidrule(lr){6-7} \cmidrule(l){8-10}
& \textbf{SFT} & \textbf{GRPO} & & & \textbf{Suc. (\%)} & \textbf{Sub-Suc. (\%)} & \textbf{Ung. (\%)} & \textbf{Sub-G. (\%)} & \textbf{Sub-Sol. (\%)} \\ 
\midrule
Base Model & $\times$ & $\times$ & $\times$ & $\times$ & 3.0 & 29.7 & 7.5 & 2.5 & 28.4 \\
SFT Stage 1 Only & \checkmark & $\times$ & \checkmark & $\times$ & 3.0 & 23.7 & 7.5 & 2.5 & 20.0 \\
GRPO Stage 1 Only & $\times$ & \checkmark & \checkmark & $\times$ & 9.1 & 33.8 & 12.5 & 5.0 & 35.2 \\
GRPO Stage 2 Only & $\times$ & \checkmark & $\times$ & \checkmark & 9.1 & 30.0 & 10.0 & 7.5 & 30.6 \\
\midrule
\textbf{Pentest-R1} & $\times$ & \checkmark & \checkmark & \checkmark & \textbf{24.2} & \textbf{33.4} & \textbf{15.0} & \textbf{15.0} & \textbf{42.8} \\ 
\bottomrule
\end{tabular}
}
\end{table*}

\section{Experiments}
This section presents a comprehensive evaluation of the Pentest-R1 framework. Our investigation is guided by the following research questions (RQs):

\textbf{RQ1:}  How does Pentest-R1's performance compare against state-of-the-art (SOTA) proprietary and open-source LLMs in autonomous penetration testing?

\textbf{RQ2:} How do the individual reinforcement learning stages (offline and online) contribute to the overall efficacy of Pentest-R1?

\textbf{RQ3:} What is the interplay between token consumption, explicit reasoning via Chain-of-Thought (CoT), and task success in cybersecurity benchmarks?

\subsection{Experimental Setup}

\paragraph{Datasets.}
Our evaluation is grounded in two prominent and challenging benchmarks designed to rigorously assess LLMs's ability in autonomous penetration testing.
\textbf{Cybench} \cite{zhang2024cybench} features 40 professional-level Capture The Flag (CTF) tasks from recent competitions, covering diverse skills like web exploitation, cryptography, and reverse engineering.  
\textbf{AutoPenBench} \cite{gioacchini2024autopenbench} is an open-source framework with 33 tasks of increasing difficulty, ranging from basic \textit{in-vitro} scenarios to complex \textit{real-world} cases based on known vulnerabilities. Its explicit task decomposition enables fine-grained evaluation of an agent's planning and execution capabilities.

\paragraph{Baselines.}
We benchmark Pentest-R1 against a diverse set of SOTA LLMs, including leading proprietary models (e.g., GPT-4o, Claude 3.7 Sonnet, Gemini 2.5 Flash, Claude 4 Opus \cite{anthropic2025claude4}) and powerful open-source models (e.g., Llama 3.1 405B \cite{meta2024llama3}, Mixtral 8x22b \cite{mistral2024mixtral8x22b}, Qwen3-32B).  For relevant models, we evaluate performance both with and without explicit CoT reasoning.

\paragraph{Evaluation Metrics.}
Our primary metric for end-to-end task completion is pass@3, which denotes a successful attempt in at least one of three independent trials, mitigating the impact of output stochasticity.
On Cybench, performance is measured by Unguided \% (Success rate without subtask guidance), Subtask-Guided \% (Success rate with subtask guidance), and the Subtasks-Solved (\%).
On AutoPenBench, we report the overall Success Rate (\%) and the percentage of completed subtasks, Subtask-Success (\%). While subtask success provides insight into plan-following, the overall success rate is the most definitive measure, as an agent may find a valid, alternative solution path.

\paragraph{Implementation Details.}
We selected \texttt{DeepSeek-R1-0528-Qwen3-8B} as the base model for Pentest-R1 due to its strong foundational capabilities. For offline training (Stage 1), we used a batch size of 16 and 4 generation candidates per prompt, training for 2 epochs. For online training (Stage 2), the agent generated 8-turn conversational trajectories; we used a batch size of 1 and 4 generation candidates, again training for 2 epochs. Both stages utilized the AdamW optimizer with a learning rate of 5e-6, a temperature of 0.6, and a cosine learning rate scheduler with a 0.03 warmup ratio. All experiments were conducted on a Kali Linux 2023 platform using two NVIDIA H100 80G GPUs, using the unsloth library \cite{unsloth}.

\subsection{Main Results (RQ1)}
We evaluated Pentest-R1 against a suite of state-of-the-art models on two challenging benchmarks, with results summarized in Table~\ref{tab:main_results}. On AutoPenBench, Pentest-R1 attains a 24.2\% success rate, decisively outperforming all open-source competitors and surpassing even formidable proprietary models like GPT-4o (21.2\%), with its performance second only to Gemini 2.5 Flash. Furthermore, it establishes a SOTA on Cybench, achieving a 15.0\% success rate in unguided tasks that is on par with Claude 3.7 Sonnet.
Critically, these completion rates were achieved through fully autonomous, end-to-end reasoning on real-time systems, without any manual prompting or subtask guidance. 
This high degree of performance is particularly remarkable given that Pentest-R1 is fine-tuned from only an 8b model, which validates the potent efficacy of our proposed two-stage training framework. Additionally, its lightweight footprint enables rapid, cost-effective deployment at scale, making it well suited for real-world penetration-testing applications.

\subsection{Ablation Study (RQ2)}
To answer RQ2 and dissect the contribution of each component, we conducted a thorough ablation study, the results of which are presented in Table~\ref{tab:ablation}. We assessed five configurations: (1) the Base Model (DeepSeek-R1-0528-Qwen3-8B) without any RL, (2) a model with only Stage 1 Supervised Fine-Tuning (SFT), (3) with only Stage 1 GRPO training, (4) with only Stage 2 GRPO training, and (5) the full, two-stage Pentest-R1 framework.
The results unequivocally demonstrate that both RL stages are critical for optimal performance. The Base Model and the SFT-only model exhibit minimal task-solving ability, with success rates of just 3.0\% on AutoPenBench. Applying GRPO in Stage 1 provides a substantial uplift, jumping to 9.1\%, which indicates that offline RL on expert data successfully imparts core competencies. However, the most significant gains are realized only with the complete pipeline. The full Pentest-R1 framework, which combines offline knowledge acquisition with online interactive refinement, markedly outperforms all partial configurations, this synergy validates our two-stage design.

\begin{figure}[t]
    \centering
    \subfigure[Cybench]{\includegraphics[width=.49\linewidth]{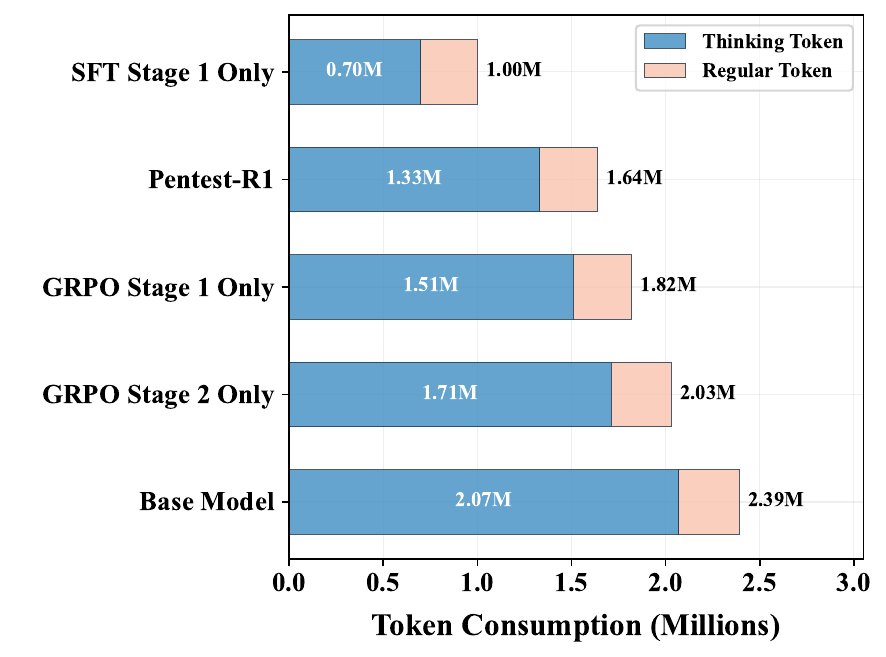}\label{fig:cybench_think}}  
    \hfill
    \subfigure[AutoPenBench]{\includegraphics[width=.49\linewidth]{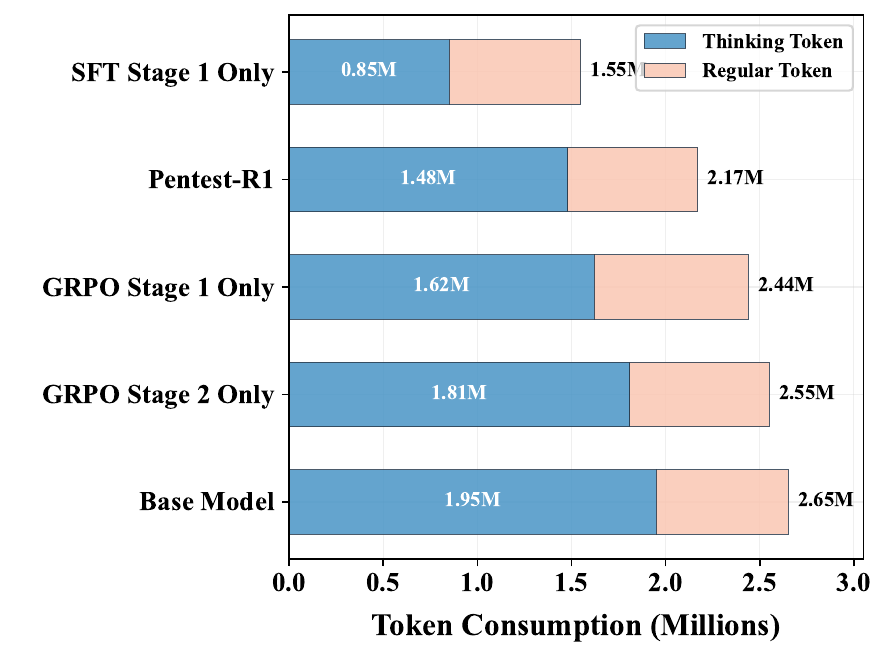}\label{fig:autopen_think}}
    \caption{Token consumption analysis. "Thinking Token" refers to tokens for Chain-of-Thought reasoning.}
    \label{figure:think}
\end{figure}

\subsection{Cost and Reasoning Analysis (RQ3)}
\begin{figure}[ht]
\centering
\includegraphics[width=0.98\linewidth]{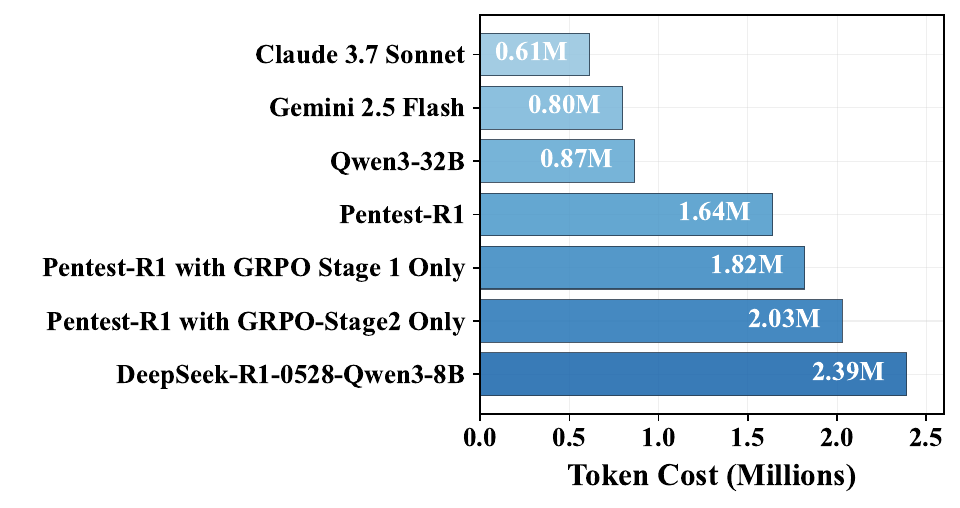}
\caption{Total token consumption on Cybench.}
\label{fig:cybench_models}
\end{figure}

We analyzed the token consumption patterns of our models, focusing on the trade-off between computational cost and performance.  As shown in Figure~\ref{figure:think}, "thinking tokens" dedicated to CoT constitute a significant computational investment, accounting for over 81\% of the total tokens on Cybench and nearly 69\% on AutoPenBench for the full Pentest-R1 model. The base model high volume of thought does not translate to success, as the base model's performance is poor. This suggests its reasoning is inefficient, often leading to unproductive or circular thought paths. The true advantage of Pentest-R1 is not that it fundamentally reduces the proportion of thinking, but that it drastically improves its quality and directness. Our two-stage reinforcement learning framework trains the model to think smarter and converge to the correct solution path more quickly. As shown in Figure~\ref{fig:cybench_models}, the fully trained Pentest-R1 agent consumes approximately 1.64 million tokens on Cybench, a 31\% reduction compared to the 2.39 million tokens used by the untrained base model. This indicates that our two-stage RL framework optimizes the base model's policy to find more direct and effective solutions. While Pentest-R1's token usage remains higher than that of highly optimized proprietary models (e.g., Claude 3.7 Sonnet at 613K and Gemini 2.5 Flash at 798K), its superior performance among open-source models validates our approach as a highly effective strategy for empowering smaller models to compete at the state-of-the-art level. Notably, the low token count of Qwen3-32B stems from a premature termination caused by an incorrect flag submission, rather than from efficient problem-solving.

\section{Conclusion}
In this paper, we introduced Pentest-R1, a novel framework designed to overcome the critical limitations of LLMs in autonomous penetration testing, including poor reasoning and error recovery. Through a specialized two-stage reinforcement learning pipeline, which first instills foundational knowledge from expert data via offline RL and then refines adaptive strategies via online RL in an interactive environment.
Our experiments demonstrate that the synergy of its two-stage reinforcement learning is essential for achieving performance that rivals and even surpasses leading proprietary models.
Furthermore, our analysis showed that while costly, explicit reasoning is crucial for success, our RL process uniquely optimizes this reasoning to be significantly more token-efficient and effective than that of the untrained base model. In future work, we will enhance our framework with multimodal capabilities—enabling the agent to interpret visual interfaces—to tackle broader and more complex cybersecurity challenges.

\bibliographystyle{IEEEtran}
\bibliography{reference}

\begin{thebibliography}{10}
\providecommand{\url}[1]{#1}
\csname url@samestyle\endcsname
\providecommand{\newblock}{\relax}
\providecommand{\bibinfo}[2]{#2}
\providecommand{\BIBentrySTDinterwordspacing}{\spaceskip=0pt\relax}
\providecommand{\BIBentryALTinterwordstretchfactor}{4}
\providecommand{\BIBentryALTinterwordspacing}{\spaceskip=\fontdimen2\font plus
\BIBentryALTinterwordstretchfactor\fontdimen3\font minus \fontdimen4\font\relax}
\providecommand{\BIBforeignlanguage}[2]{{%
\expandafter\ifx\csname l@#1\endcsname\relax
\typeout{** WARNING: IEEEtran.bst: No hyphenation pattern has been}%
\typeout{** loaded for the language `#1'. Using the pattern for}%
\typeout{** the default language instead.}%
\else
\language=\csname l@#1\endcsname
\fi
#2}}
\providecommand{\BIBdecl}{\relax}
\BIBdecl

\bibitem{metasploit2020}
O.~Valea and C.~Opri{\c{s}}a, ``Towards pentesting automation using the metasploit framework,'' in \emph{2020 IEEE 16th International Conference on Intelligent Computer Communication and Processing (ICCP)}.\hskip 1em plus 0.5em minus 0.4em\relax IEEE, 2020, pp. 171--178.

\bibitem{achiam2023gpt}
J.~Achiam, S.~Adler, S.~Agarwal, L.~Ahmad, I.~Akkaya, F.~L. Aleman, D.~Almeida, J.~Altenschmidt, S.~Altman, S.~Anadkat \emph{et~al.}, ``Gpt-4 technical report,'' \emph{arXiv preprint arXiv:2303.08774}, 2023.

\bibitem{guo2025deepseek}
D.~Guo, D.~Yang, H.~Zhang, J.~Song, R.~Zhang, R.~Xu, Q.~Zhu, S.~Ma, P.~Wang, X.~Bi \emph{et~al.}, ``Deepseek-r1: Incentivizing reasoning capability in llms via reinforcement learning,'' \emph{arXiv preprint arXiv:2501.12948}, 2025.

\bibitem{comanici2025gemini}
G.~Comanici, E.~Bieber, M.~Schaekermann, I.~Pasupat, N.~Sachdeva, I.~Dhillon, M.~Blistein, O.~Ram, D.~Zhang, E.~Rosen \emph{et~al.}, ``Gemini 2.5: Pushing the frontier with advanced reasoning, multimodality, long context, and next generation agentic capabilities,'' \emph{arXiv preprint arXiv:2507.06261}, 2025.

\bibitem{deng2024pentestgpt}
G.~Deng, Y.~Liu, V.~Mayoral-Vilches, P.~Liu, Y.~Li, Y.~Xu, T.~Zhang, Y.~Liu, M.~Pinzger, and S.~Rass, ``$\{$PentestGPT$\}$: Evaluating and harnessing large language models for automated penetration testing,'' in \emph{33rd USENIX Security Symposium (USENIX Security 24)}, 2024, pp. 847--864.

\bibitem{kong2025vulnbot}
H.~Kong, D.~Hu, J.~Ge, L.~Li, T.~Li, and B.~Wu, ``Vulnbot: Autonomous penetration testing for a multi-agent collaborative framework,'' \emph{arXiv preprint arXiv:2501.13411}, 2025.

\bibitem{wei2022chain}
J.~Wei, X.~Wang, D.~Schuurmans, M.~Bosma, F.~Xia, E.~Chi, Q.~V. Le, D.~Zhou \emph{et~al.}, ``Chain-of-thought prompting elicits reasoning in large language models,'' \emph{Advances in neural information processing systems}, vol.~35, pp. 24\,824--24\,837, 2022.

\bibitem{yang2023intercode}
J.~Yang, A.~Prabhakar, K.~Narasimhan, and S.~Yao, ``Intercode: Standardizing and benchmarking interactive coding with execution feedback,'' \emph{Advances in Neural Information Processing Systems}, vol.~36, pp. 23\,826--23\,854, 2023.

\bibitem{foley2025apirl}
M.~Foley and S.~Maffeis, ``Apirl: Deep reinforcement learning for rest api fuzzing,'' in \emph{Proceedings of the AAAI Conference on Artificial Intelligence}, vol.~39, no.~1, 2025, pp. 191--199.

\bibitem{xiong2025minimalist}
W.~Xiong, J.~Yao, Y.~Xu, B.~Pang, L.~Wang, D.~Sahoo, J.~Li, N.~Jiang, T.~Zhang, C.~Xiong \emph{et~al.}, ``A minimalist approach to llm reasoning: from rejection sampling to reinforce,'' \emph{arXiv preprint arXiv:2504.11343}, 2025.

\bibitem{mroueh2025reinforcement}
Y.~Mroueh, ``Reinforcement learning with verifiable rewards: Grpo's effective loss, dynamics, and success amplification,'' \emph{arXiv preprint arXiv:2503.06639}, 2025.

\bibitem{shao2024deepseekmath}
Z.~Shao, P.~Wang, Q.~Zhu, R.~Xu, J.~Song, X.~Bi, H.~Zhang, M.~Zhang, Y.~Li, Y.~Wu \emph{et~al.}, ``Deepseekmath: Pushing the limits of mathematical reasoning in open language models,'' \emph{arXiv preprint arXiv:2402.03300}, 2024.

\bibitem{hu2022lora}
E.~J. Hu, Y.~Shen, P.~Wallis, Z.~Allen-Zhu, Y.~Li, S.~Wang, L.~Wang, W.~Chen \emph{et~al.}, ``Lora: Low-rank adaptation of large language models.'' \emph{ICLR}, vol.~1, no.~2, p.~3, 2022.

\bibitem{zhang2024cybench}
A.~K. Zhang, N.~Perry, R.~Dulepet, J.~Ji, C.~Menders, J.~W. Lin, E.~Jones, G.~Hussein, S.~Liu, D.~Jasper \emph{et~al.}, ``Cybench: A framework for evaluating cybersecurity capabilities and risks of language models,'' \emph{arXiv preprint arXiv:2408.08926}, 2024.

\bibitem{gioacchini2024autopenbench}
L.~Gioacchini, M.~Mellia, I.~Drago, A.~Delsanto, G.~Siracusano, and R.~Bifulco, ``Autopenbench: Benchmarking generative agents for penetration testing,'' \emph{arXiv preprint arXiv:2410.03225}, 2024.

\bibitem{moscovich2020autosploit}
N.~Moscovich, R.~Bitton, Y.~Mallah, M.~Inokuchi, T.~Yagyu, M.~Kalech, Y.~Elovici, and A.~Shabtai, ``Autosploit: A fully automated framework for evaluating the exploitability of security vulnerabilities,'' \emph{arXiv preprint arXiv:2007.00059}, 2020.

\bibitem{happe2025can}
A.~Happe and J.~Cito, ``Can llms hack enterprise networks? autonomous assumed breach penetration-testing active directory networks,'' \emph{arXiv preprint arXiv:2502.04227}, 2025.

\bibitem{mayoral2025cai}
V.~Mayoral-Vilches, L.~J. Navarrete-Lozano, M.~Sanz-G{\'o}mez, L.~S. Espejo, M.~Crespo-{\'A}lvarez, F.~Oca-Gonzalez, F.~Balassone, A.~Glera-Pic{\'o}n, U.~Ayucar-Carbajo, J.~A. Ruiz-Alcalde \emph{et~al.}, ``Cai: An open, bug bounty-ready cybersecurity ai,'' \emph{arXiv preprint arXiv:2504.06017}, 2025.

\bibitem{xu2024autoattacker}
J.~Xu, J.~W. Stokes, G.~McDonald, X.~Bai, D.~Marshall, S.~Wang, A.~Swaminathan, and Z.~Li, ``Autoattacker: A large language model guided system to implement automatic cyber-attacks,'' \emph{arXiv preprint arXiv:2403.01038}, 2024.

\bibitem{NEURIPS2024_69d97a64}
\BIBentryALTinterwordspacing
M.~Shao, S.~Jancheska, M.~Udeshi, B.~Dolan-Gavitt, H.~Xi, K.~Milner, B.~Chen, M.~Yin, S.~Garg, P.~Krishnamurthy, F.~Khorrami, R.~Karri, and M.~Shafique, ``Nyu ctf bench: A scalable open-source benchmark dataset for evaluating llms in offensive security,'' in \emph{Advances in Neural Information Processing Systems}, A.~Globerson, L.~Mackey, D.~Belgrave, A.~Fan, U.~Paquet, J.~Tomczak, and C.~Zhang, Eds., vol.~37.\hskip 1em plus 0.5em minus 0.4em\relax Curran Associates, Inc., 2024, pp. 57\,472--57\,498. [Online]. Available: \url{https://proceedings.neurips.cc/paper_files/paper/2024/file/69d97a6493fbf016fff0a751f253ad18-Paper-Datasets_and_Benchmarks_Track.pdf}
\BIBentrySTDinterwordspacing

\bibitem{el2025competitive}
A.~El-Kishky, A.~Wei, A.~Saraiva, B.~Minaiev, D.~Selsam, D.~Dohan, F.~Song, H.~Lightman, I.~Clavera, J.~Pachocki \emph{et~al.}, ``Competitive programming with large reasoning models,'' \emph{arXiv preprint arXiv:2502.06807}, 2025.

\bibitem{jaech2024openai}
A.~Jaech, A.~Kalai, A.~Lerer, A.~Richardson, A.~El-Kishky, A.~Low, A.~Helyar, A.~Madry, A.~Beutel, A.~Carney \emph{et~al.}, ``Openai o1 system card,'' \emph{arXiv preprint arXiv:2412.16720}, 2024.

\bibitem{hurst2024gpt}
A.~Hurst, A.~Lerer, A.~P. Goucher, A.~Perelman, A.~Ramesh, A.~Clark, A.~Ostrow, A.~Welihinda, A.~Hayes, A.~Radford \emph{et~al.}, ``Gpt-4o system card,'' \emph{arXiv preprint arXiv:2410.21276}, 2024.

\bibitem{zhang2024llama}
D.~Zhang, J.~Wu, J.~Lei, T.~Che, J.~Li, T.~Xie, X.~Huang, S.~Zhang, M.~Pavone, Y.~Li \emph{et~al.}, ``Llama-berry: Pairwise optimization for o1-like olympiad-level mathematical reasoning,'' \emph{arXiv preprint arXiv:2410.02884}, 2024.

\bibitem{xu2024llava}
G.~Xu, P.~Jin, L.~Hao, Y.~Song, L.~Sun, and L.~Yuan, ``Llava-o1: Let vision language models reason step-by-step,'' \emph{arXiv preprint arXiv:2411.10440}, 2024.

\bibitem{zhang2024o1}
Y.~Zhang, S.~Wu, Y.~Yang, J.~Shu, J.~Xiao, C.~Kong, and J.~Sang, ``o1-coder: an o1 replication for coding,'' \emph{arXiv preprint arXiv:2412.00154}, 2024.

\bibitem{zhao2024marco}
Y.~Zhao, H.~Yin, B.~Zeng, H.~Wang, T.~Shi, C.~Lyu, L.~Wang, W.~Luo, and K.~Zhang, ``Marco-o1: Towards open reasoning models for open-ended solutions,'' \emph{arXiv preprint arXiv:2411.14405}, 2024.

\bibitem{qwq32b}
\BIBentryALTinterwordspacing
Q.~Team, ``Qwq-32b: Embracing the power of reinforcement learning,'' March 2025. [Online]. Available: \url{https://qwenlm.github.io/blog/qwq-32b/}
\BIBentrySTDinterwordspacing

\bibitem{team2023gemini}
G.~Team, R.~Anil, S.~Borgeaud, J.-B. Alayrac, J.~Yu, R.~Soricut, J.~Schalkwyk, A.~M. Dai, A.~Hauth, K.~Millican \emph{et~al.}, ``Gemini: a family of highly capable multimodal models,'' \emph{arXiv preprint arXiv:2312.11805}, 2023.

\bibitem{deng2024novice}
Z.~Deng, Z.~Dou, Y.~Zhu, J.-R. Wen, R.~Xiong, M.~Wang, and W.~Chen, ``From novice to expert: Llm agent policy optimization via step-wise reinforcement learning,'' \emph{arXiv preprint arXiv:2411.03817}, 2024.

\bibitem{feng2025group}
L.~Feng, Z.~Xue, T.~Liu, and B.~An, ``Group-in-group policy optimization for llm agent training,'' \emph{arXiv preprint arXiv:2505.10978}, 2025.

\bibitem{Agent-R1}
\BIBentryALTinterwordspacing
O.~Jie, Y.~Ruiran, L.~Yucong, C.~Mingyue, L.~Qi, L.~Zirui, Y.~Shuo, and D.~Wang, ``Training powerful llm agents with end-to-end reinforcement learning,'' GitHub, 2025. [Online]. Available: \url{https://github.com/0russwest0/Agent-R1}
\BIBentrySTDinterwordspacing

\bibitem{jin2025search}
B.~Jin, H.~Zeng, Z.~Yue, J.~Yoon, S.~Arik, D.~Wang, H.~Zamani, and J.~Han, ``Search-r1: Training llms to reason and leverage search engines with reinforcement learning,'' \emph{arXiv preprint arXiv:2503.09516}, 2025.

\bibitem{wang2025ragen}
Z.~Wang, K.~Wang, Q.~Wang, P.~Zhang, L.~Li, Z.~Yang, X.~Jin, K.~Yu, M.~N. Nguyen, L.~Liu \emph{et~al.}, ``Ragen: Understanding self-evolution in llm agents via multi-turn reinforcement learning,'' \emph{arXiv preprint arXiv:2504.20073}, 2025.

\bibitem{feng2025doctoragent}
Y.~Feng, J.~Wang, L.~Zhou, and Y.~Li, ``Doctoragent-rl: A multi-agent collaborative reinforcement learning system for multi-turn clinical dialogue,'' \emph{arXiv preprint arXiv:2505.19630}, 2025.

\bibitem{yang2025qwen3}
A.~Yang, A.~Li, B.~Yang, B.~Zhang, B.~Hui, B.~Zheng, B.~Yu, C.~Gao, C.~Huang, C.~Lv \emph{et~al.}, ``Qwen3 technical report,'' \emph{arXiv preprint arXiv:2505.09388}, 2025.

\bibitem{claude3.7sonnet}
\BIBentryALTinterwordspacing
Anthropic. (2025) Claude 3.7 sonnet. [Online]. Available: \url{https://www.anthropic.com/news/claude-3-7-sonnet}
\BIBentrySTDinterwordspacing

\bibitem{kali}
Kali, ``The most advanced penetration testing distribution,'' 2024, \url{https://www.kali.org/}.

\bibitem{hackthebox}
{Hack The Box}, ``Hack the box: A hacking playground,'' \url{https://www.hackthebox.com}, 2025, accessed: [2025-07-01].

\bibitem{vulnhub}
{VulnHub}, ``Vulnhub: Vulnerable by design,'' \url{https://www.vulnhub.com}, 2025, accessed: [2025-07-01].

\bibitem{anthropic2025claude4}
{Anthropic}, ``Claude\,4 opus,'' \url{https://www.anthropic.com/news/claude-4}, 2025.

\bibitem{meta2024llama3}
\BIBentryALTinterwordspacing
M.~AI, ``Introducing meta llama 3: The most capable openly available llm to date,'' April 2024. [Online]. Available: \url{https://ai.meta.com/blog/meta-llama-3/}
\BIBentrySTDinterwordspacing

\bibitem{mistral2024mixtral8x22b}
\BIBentryALTinterwordspacing
------, ``Mixtral 8x22b,'' April 2024. [Online]. Available: \url{https://mistral.ai/news/mixtral-8x22b/}
\BIBentrySTDinterwordspacing

\bibitem{unsloth}
\BIBentryALTinterwordspacing
M.~H. Daniel~Han and U.~team, ``Unsloth,'' 2023. [Online]. Available: \url{http://github.com/unslothai/unsloth}
\BIBentrySTDinterwordspacing

\end{thebibliography}
\end{document}